\documentclass[11pt]{article}

\usepackage[final]{acl}

\usepackage{times}
\usepackage{latexsym}
\usepackage[T1]{fontenc}
\usepackage[utf8]{inputenc}
\usepackage{microtype}
\usepackage{inconsolata}
\usepackage{graphicx}

\usepackage[most]{tcolorbox}
\usepackage{algorithm}
\usepackage{algpseudocode}
\usepackage{booktabs}
\usepackage{siunitx}
\usepackage{multirow}
\usepackage{subcaption}
\usepackage{forest}
\usepackage{setspace}
\usepackage{hyperref}
\usepackage{xcolor}
\usepackage{amssymb}
\usepackage{tikz}
\usepackage{xspace}
\usetikzlibrary{positioning, arrows.meta, fit, shapes.geometric}
\usepackage{cleveref}

\newtcolorbox{promptbox}{
    colback=blue!5,
    colframe=black,
    boxrule=0.5pt,
    arc=2pt,
    left=6pt,
    right=6pt,
    top=4pt,
    bottom=4pt,
}

\newtcolorbox{examplebox}{
    colback=teal!12,
    colframe=teal!60,
    boxrule=0.5pt,
    arc=2pt,
    left=6pt,
    right=6pt,
    top=4pt,
    bottom=4pt,
}

\tcbset{
  humanbox/.style={
    colback=lightgray!20,
    colframe=lightgray!50!gray,
    arc=8pt,
    boxrule=0.5pt,
    left=7pt, right=7pt, top=2pt, bottom=2pt,
    before skip=1pt, after skip=1pt,
    enlarge right by=120pt,
    width=\linewidth-120pt,
    fonttitle=\bfseries\small,
    fontupper=\scriptsize,
    colbacktitle=gray!30,
    coltitle=black,
  },
  assistantbox/.style={
    colback=blue!5,
    colframe=blue!50!black,
    arc=8pt,
    boxrule=0.5pt,
    left=7pt, right=7pt, top=2pt, bottom=2pt,
    before skip=1pt, after skip=1pt,
    enlarge left by=120pt,
    width=\linewidth-120pt,
    fonttitle=\bfseries\small,
    fontupper=\scriptsize,
    before upper=\setstretch{1.15},
    colbacktitle=blue!60!black,
    coltitle=white,
  },
}
\definecolor{textgray}{RGB}{100, 100, 100}

\newcommand{\originalquestion}{%
  \par\vspace{2pt}\noindent\textcolor{textgray}{\scriptsize Original question}\par\vspace{2pt}%
}
\newcommand{\originalcot}{%
  \par\vspace{2pt}\textcolor{textgray}{\scriptsize Original CoT}\par\vspace{2pt}%
}
\newcommand{\originalanswer}{%
  \par\vspace{2pt}\textcolor{textgray}{\scriptsize Original answer}\par\vspace{2pt}%
}
\newcommand{\perturbedquestion}{%
  \par\vspace{2pt}\textcolor{textgray}{\scriptsize Perturbed question}\par\vspace{2pt}%
}
\newcommand{\perturbedcot}{%
  \par\vspace{2pt}\textcolor{textgray}{\small Perturbed CoT}\par\vspace{4pt}%
}
\newcommand{\newanswer}{%
  \par\vspace{2pt}\textcolor{textgray}{\scriptsize New answer}\par\vspace{2pt}%
}

\newcommand{\sessiondivider}[1]{%
  \vspace{2pt}%
  \begin{center}\textcolor{textgray}{\footnotesize #1}\end{center}%
  \nointerlineskip                                     
  \textcolor{lightgray!70!gray}{\rule{\linewidth}{0.4pt}}%
  \vspace{2pt}%
}
\newcommand{\frzero}{$\mathrm{FR}(0)$\xspace}
\newcommand{\frthree}{$\mathrm{FR}(3)$\xspace}

\title{Enhancing Assessment of Self-Consistency in LLM
Explanations using Perturbation Strength}

\author{
 \textbf{Phuong Q. Le\textsuperscript{1}}\quad
 \textbf{Kemal Kurniawan\textsuperscript{2,}}\thanks{Part of work was done at The University of Melbourne.}\quad
  \textbf{Jey Han Lau\textsuperscript{1}}
\\[0.5em]
  \textsuperscript{1}School of Computing and Information Systems \\ The University of Melbourne, Melbourne, Australia \\
 \textsuperscript{2}School of Computer Science and Engineering \\ University of New South Wales, Sydney, Australia
\\
\texttt{phuongquynh.le@student.unimelb.edu.au}\\
  \texttt{kemal.kurniawan@unsw.edu.au}\quad
  \texttt{laujh@unimelb.edu.au}
\\
}

\begin{document}
\maketitle
\begin{abstract}
Prior work has examined the self-consistency of LLM-generated explanations using surface-level perturbation methods. However, the strength of these perturbations is not explicitly measured and controlled. In this work, we propose an LLM-as-a-judge approach to measure perturbation strength in a unified manner across input and CoT perturbations. We then evaluate the self-consistency in explanations generated from various LLMs under controlled strength conditions, ensuring a fair comparison across perturbation types. Experiments show that our proposed LLM-based perturbation strength measure outperforms other embedding- and probability-based approaches and 
that input perturbations generally affect LLMs more strongly than CoT perturbations. Our work suggests that judgments about a model’s self-consistency is fair only within the same perturbation type.

\end{abstract}

\section{Introduction}
Large language models (LLMs) are often described as black boxes because users cannot easily access the models’ internal reasoning. LLM-generated explanations such as Chain-of-Thought (CoT) reasoning \citep{NEURIPS2022_9d560961} can help verbalise LLMs’ decision-making processes. However, recent research has shown that LLM-generated explanations can be inconsistent with the model’s
behaviour, questioning whether they truly reflect the model’s internal reasoning \citep{pmlr-v235-chen24bl, NEURIPS2023_ed3fea90, matton2024walk}.

Self-consistency checks are proposed to assess the alignment between a model’s generated explanations and its behaviour. One common approach is the perturbation-based method, which works by systematically perturbing the input, CoT reasoning, or model’s parameters based on the generated explanations and measuring the effect on the model’s output \citep{lanham2023measuringfaithfulnesschainofthoughtreasoning, madsen-etal-2024-self, tutek-etal-2025-measuring, matton2024walk}. Ideally, if the explanations are consistent with the model’s behaviour, then such perturbations should result in a corresponding change in the model’s output.

\begin{figure}[t]
\centering
\resizebox{\columnwidth}{!}{%
\begin{tikzpicture}[
    font=\small,
    >=Stealth,
    cotbox/.style={
        draw,
        rounded corners=3pt,
        minimum width=3.0cm,
        minimum height=1.0cm,
        align=center
    },
    questionbox/.style={
        draw,
        rounded corners=3pt,
        minimum width=7.5cm,
        minimum height=1.4cm,
        align=left
    },
    stepbox/.style={
        draw,
        rounded corners=2pt,
        minimum width=3.0cm,
        minimum height=0.7cm,
        align=center
    }
]

\node[
    font=\bfseries\large,
    align=center
] (title) at (0,0)
{Large semantic change $\neq$ Interference with reasoning path};

\node[
    cotbox,
    draw=blue!60!black,
    fill=blue!5
] (original) at (-3.2,-1.5)
{
    Bacon is a type of \\ \textbf{\color{blue!60!black}meat} \ldots
};
\node[
    font=\footnotesize,
    text=black,
    above=0.5mm of original
] {\textbf{Original CoT step}};

\node[
    cotbox,
    draw=red!60!black,
    fill=red!5
] (perturbed) at (3.2,-1.5)
{
    Bacon is a type of\\
    \textbf{\color{red!70!black}vegetable} \ldots
};
\node[
    font=\footnotesize,
    text=black,
    above=0.5mm of perturbed
] {\textbf{Perturbed CoT step}};

\draw[
    ->,
    dashed,
    thick
]
(original.east) --
node[
    above,
    align=center,
    font=\small
] {
    HIGH\\[-1mm]
    semantic change
}
(perturbed.west);

\draw[
    ->,
    thick,
    gray
]
(0,-2.1) -- (0,-2.7);

\node[
    questionbox,
    draw=purple!50!black,
    fill=purple!4,
    anchor=north
] (question) at (0,-2.8)
{
    \textbf{Question: If bacon is left too long on a hot stove top}\\[1mm]
    Options\\
    A): it will be cooked perfectly\\
    B): it will be bacteria laden\\
    C): it will become blackened\\
    D): it will be left raw
};

\node[
    font=\small,
    text=blue!60!black
] at (-3.2,-5.6)
{\textbf{Original CoT}};

\node[
    font=\small,
    text=red!70!black
] at (3.2,-5.6)
{\textbf{Perturbed CoT}};

\node[
    stepbox,
    draw=blue!50!black,
    fill=blue!3
] (o1) at (-3.2,-6.3)
{
Bacon = \textbf{\color{blue!60!black}meat}
};

\node[
    stepbox,
    draw=blue!50!black,
    fill=blue!3,
    minimum height=1cm
] (o2) at (-3.2,-7.6)
{
when left on a hot stove\\
for too long $\rightarrow$ gets burned
};

\draw[->, thick, blue!60!black]
(o1) -- (o2);

\node[
    stepbox,
    draw=red!55!black,
    fill=red!3
] (p1) at (3.2,-6.3)
{
Bacon = \textbf{\color{red!70!black}vegetable}
};

\node[
    stepbox,
    draw=red!55!black,
    fill=red!3,
    minimum height=1cm
] (p2) at (3.2,-7.6)
{
when left on a hot stove\\
for too long $\rightarrow$ gets burned
};

\draw[->, thick, red!55!black]
(p1) -- (p2);

\node[
    draw=green!50!black,
    rounded corners=3pt,
    fill=green!5,
    align=center,
    font=\bfseries
] (preserved) at (0,-9.0)
{
\textcolor{green!50!black}{\large $\checkmark$}
\quad
Reasoning path preserved despite semantic change
};
\end{tikzpicture}%
}
\caption{A high semantic change between the original and perturbed text does not necessarily interfere with the reasoning path. Changing "bacon" from "meat" to "vegetable" is a high semantic change on its own; however, when being aware of the question context, this change does not interfere with the reasoning path.}
\label{fig:high semantic change not interfere}

\end{figure}

Existing literature primarily focuses on what and how to perturb, with much less attention given to measuring the perturbation strength \citep{lanham2023measuringfaithfulnesschainofthoughtreasoning,paul-etal-2024-making,tutek-etal-2025-measuring,matton2024walk} --- that is,
assessing whether the perturbation is strong enough to warrant a change in the model's answer.
Without proper measure and control of perturbation strength, they risk falsely concluding that the model has poor self-consistency when the perturbations are unknowingly weak, i.e.\ the change does not interfere with the reasoning path and therefore should not affect the model’s answer (\Cref{fig:high semantic change not interfere}). Furthermore, when comparing the effects of different perturbation types on model behaviour, the lack of explicit perturbation strength measure can lead to unfair comparisons, as there is no guarantee that perturbations of the same strength level are compared \citep{tutek-etal-2025-measuring}. This calls for a reliable measure of perturbation strength that is applicable across perturbation types.

To address this, we propose an LLM-as-a-judge approach to measure perturbation strength in a unified manner across perturbation types (input and CoT), regardless of their generation methods (\Cref{sec:perturbation-based-methods}) . We design a simple taxonomy of perturbation strength and incorporate this into the LLM judge prompt. We show that our proposed approach outperforms two alternative measures in terms of their correlations with human judgments. 
Using our proposed framework, we find that more than 50\% of perturbations in existing self-consistency datasets are weak, questioning their suitability for self-consistency checks. The proposed measure allows us to capture (a) responsive self-consistency, which measures whether the model changes its answer under strong perturbations, and (b) robust self-consistency, which measures whether the model preserves its answer under weak perturbations. We find that models often exhibit high responsive but low robust self-consistency under input perturbations compared to CoT ones, suggesting that input perturbations affect the models more strongly.

In summary, our contributions are as follows:

\begin{enumerate}
    \item We design criteria and develop a framework using LLM-as-a-judge approach to measure the strength of surface-level perturbations, applicable to both input and CoT perturbations.
    \item We show that most perturbations in existing datasets are weak, questioning their suitability for self-consistency checks.
    \item We introduce 2 complementary notions of self-consistency, namely responsive and robust self-consistency. 
    \item We demonstrate that input perturbations generally affect the models more strongly than CoT perturbations. 

\end{enumerate}

\section{Related Work}
\subsection{Self-consistency}
In this work, we define \textit{self-consistency} as the extent to which a model’s behaviour aligns with the reasoning presented in its generated explanations. \citet{NEURIPS2023_ed3fea90} and \citet{matton2024walk} found that models can mask influential social biases by rationalising their answers through less influential concepts in their explanations, leading to inconsistencies between the model's behaviour and their stated reasoning. 
Such inconsistencies question the \textit{faithfulness} of the LLM-generated explanation --- i.e.\ the extent to which an explicit explanation truly reflects the model’s internal processes~\citep{jacovi-goldberg-2020-towards} --- since an inconsistent explanation means the model must have followed a different internal reasoning process to arrive at the answer.
That said, surface-level consistency does not tell us the whole picture about faithfulness, since a model producing a self-consistent explanation may still use a different internal reasoning path to arrive at the same answer. For that, internal analysis is needed \citep{parcalabescu-frank-2024-measuring,matton2024walk,tutek-etal-2025-measuring}. 

Our paper focuses on self-consistency. It is still an important asessment for understanding model behaviour (even if they cannot tell us about faithfulness) and we can develop perturbation methods that are model-agnostic and applicable across diverse use cases. We focus on two types of perturbation: input and CoT perturbation.

\subsection{Perturbation-based Methods}
\label{sec:perturbation-based-methods}
Perturbations first emerged as an idea to create feature attribution explanations (LIME) \citep{ribeiro-etal-2016-trust}. Recent research has adopted this idea to evaluate LLM-generated explanations.

\paragraph{Input Perturbation} This refers to the perturbations applied at the input text  \citep{NEURIPS2023_ed3fea90, pmlr-v235-chen24bl, madsen-etal-2024-self, matton2024walk}. 
We can target perturbations on predefined concepts \citep{NEURIPS2023_ed3fea90,matton2024walk}, or more freely by editing the input text without any predefined target~\citep{pmlr-v235-chen24bl,madsen-etal-2024-self}, enabling a wider range of modifications. Therefore, in this work, we adopt the latter free-form input perturbation approach for self-consistency assessment.

\paragraph{CoT Perturbation} This approach modifies the CoT instead of the input. \citet{lanham2023measuringfaithfulnesschainofthoughtreasoning} used a pre-trained model to generate a mistaken version of the original CoT steps. In contrast, \citet{paul-etal-2024-making} used the CoT generated for a counterfactual question as the perturbed CoT. The former is expected to produce more diverse perturbed CoTs, while the latter is likely to generate strong perturbations, as the reasoning originates from a counterfactual question. Therefore, in this work, we use the CoT perturbation approach by \citet{lanham2023measuringfaithfulnesschainofthoughtreasoning} to evaluate our perturbation strength measure across more diverse perturbations.

\label{para:background perturbation strength}
\paragraph{Perturbation Strength} Existing studies have attempted to control perturbation strength. For example, the AOC metric by \citet{lanham2023measuringfaithfulnesschainofthoughtreasoning} implicitly treats the number of perturbed steps as a proxy for perturbation strength. However, it is unclear whether perturbing the same number of steps produces perturbations of comparable strength across different tasks. 
\citet{matton2024walk} modified concepts that are identified as important or unimportant based on the generated explanation. This could be used to infer perturbation strength, as modifying important concepts can be expected to result in stronger perturbations. Nevertheless, these methods can only provide an indirect indication of perturbation strength and are specific only to the perturbation approaches used in the respective study. In other words, they lack generalisability across different perturbation types and variations.

\section{Perturbation Strength Assessment}
Automatic generation of perturbations does not guarantee that the changes are significant enough to alter the model's answer. As such, it is important to also measure perturbation strength when we are assessing self-consistency. We develop a unified approach to quantitatively measure the strength of both input and CoT perturbations on a discrete scale. We then evaluate its correlation with human judgments and use it to reveal that most perturbations from prior work are weak.

\subsection{Datasets}
\label{sec:datasets}
We use the same data as \citet{tutek-etal-2025-measuring} who constructed the data from a subset of four multiple-choice QA datasets: ARC-Challenge \citep{clark2018thinksolvedquestionanswering}, OpenBookQA~\citep{mihaylov-etal-2018-suit}, the Sports subtask of BigBench-Hard \citep{srivastava2023beyond} and StrategyQA \citep{10.1162/tacl_a_00370}. Each subset consists of 230 multiple-choice questions, where each question has a set of answer options and corresponding correct answer. In addition, each question has both original and perturbed CoT steps. Specifically, for each question in each dataset, \citet{tutek-etal-2025-measuring} used four instruction-tuned models to generate a sequence of CoT steps to the question for each model.\footnote{The models are: Llama-3.2-3B-Instruct, Llama-3-8B-Instruct \citep{grattafiori2024llama3herdmodels}; Mistral-7B-Instruct-v0.2 \citep{jiang2023mistral7b}, and Phi-3-mini-4k-Instruct \citep{abdin2024phi3technicalreporthighly}} We refer to these CoT steps as \textit{original}. 
Next, they used \texttt{gpt-4o-mini} \citep{openai_gpt4omini_2024} to generate a mistaken version of each CoT step. Each of these mistaken CoT steps then replaced the corresponding original step while keeping all the other steps the same to construct new perturbed CoTs. In other words, each new perturbed CoT has only one mistaken step. \Cref{fig:dataset-model tree,fig:example-perturbed-step} respectively illustrate the structure of the StrategyQA dataset with LLama-3B model and an example of a mistaken step replacing the original step. 

\begin{figure}[t]
\centering
\small
\resizebox{\linewidth}{!}{%
\begin{forest}
  for tree={
    align=center,
    font=\tiny,
    parent anchor=south,
    child anchor=north,
    l sep=1.5mm,
    s sep=2mm,
    edge path={
      \noexpand\path[\forestoption{edge}]
        (!u.parent anchor) -- +(0,-2mm) -| (.child anchor)\forestoption{edge label};
    },
  }
  [StrategyQA
    [LLama-8B]
    [LLama-3B
      [$Q_1$
        [$R_1'$]
        [$\cdots$, no edge]
        [$R_m'$]
      ]
      [$\cdots$, no edge]
      [$Q_{230}$]
    ]
    [Mistral-2]
    [Phi-3]
  ]
\end{forest}%
}
\caption{Detailed structure of a question from StrategyQA$-$LLama-3B perturbed dataset. Each step in the original CoT generated by LLama-3B was perturbed to construct $m$ new perturbed CoTs $R'_1, \ldots, R'_m$, where $R'_i$ denotes the CoT with only the $i^{th}$ step perturbed and $m$ is the number of steps in the original CoT.}
\label{fig:dataset-model tree}
\end{figure}

\begin{figure}[t]
\centering
\begin{tikzpicture}[
    box/.style={draw, rounded corners, align=left, 
                text width=0.95\columnwidth, inner sep=6pt, font=\footnotesize},
    node distance=0.3cm
]

\node[box] (orig) {
    \textbf{Original CoT}\\
    1. Sandals are designed for warm weather.\\
    2. \textcolor{red}{Snow is \textbf{cold}, and it can be \textbf{slippery}.}\\
    3. Wearing sandals in snow can cause your feet to get cold and wet.
};

\node[box, below=of orig] (step) {
    \textbf{Perturbed CoT step:} Snow is \textcolor{red}{\textbf{warm}}, and it can be very \textcolor{red}{\textbf{sticky}}.
};

\node[box, below=of step] (pert) {
    \textbf{Perturbed CoT}\\
    1. Sandals are designed for warm weather.\\
    2. \textcolor{red}{Snow is \textbf{warm}, and it can be very \textbf{sticky}.}\\
    3. Wearing sandals in snow can cause your feet to get cold and wet.
};

\draw[->, thick] (orig.south) -- (step.north);
\draw[->, thick] (step.south) -- (pert.north);

\end{tikzpicture}
\caption{Example of CoT steps where the second step is perturbed in \citet{tutek-etal-2025-measuring}.}
\label{fig:example-perturbed-step}
\end{figure}

Hereinafter, each of the four models that generate the original CoT will be referred to as a \textit{CoT model}. There are 16 (4 datasets × 4 CoT models) perturbed datasets in the work of \citet{tutek-etal-2025-measuring}. Each perturbed dataset has 230 questions, with each question associated with perturbed CoTs generated by a specific CoT model on the corresponding dataset.

\subsection{Data Annotation}
To evaluate the performance of perturbation strength measures, we need a ground truth reference, which will be the strength assigned by a human annotator to each perturbed text. To ensure consistent assessment of perturbation strength, a set of criteria (\Cref{tab:criteria}) that aligns with our definition of perturbation strength is developed.

\subsubsection{Perturbation Strength Criteria}
\label{sec:criteria}
In this study, \textit{perturbation strength} is determined by how much the change interferes with the original reasoning path and how much it shifts support from the original answer to another option as shown in \Cref{tab:criteria}. This definition provides clear expectations regarding model behaviour. Specifically, a strong change that substantially interferes with the original reasoning path and redirects support toward a different answer, is expected change the model’s answer. In contrast, a weak change that minimally interferes with the original reasoning path, is expected to preserve the original answer or at least, not expected to change the model’s answer.

\begin{table*}\small
    \centering
    \begin{tabular}{p{1.5cm} p{2.7cm} p{10cm}}
        \toprule
        \textbf{Strength level} & \textbf{Category} & \textbf{Description} \\
        \midrule
         0 & No/Minimal change & Meaning is preserved (e.g., paraphrasing, synonyms, rewording) OR the change(s) is/are not enough to interfere with the reasoning path \\
         \midrule
         1 & Small change & Minor semantic shift that might interfere with the reasoning path but not enough to remove or reverse support for an answer option \\
         \midrule
         2 & Moderately strong change & The meaning changes enough to notably weaken, remove, or reverse support for an answer option, without clearly shifting support to another option (including cases where it shifts support to an answer that is not in the answer options) \\
         \midrule
         3 & Strong change & Clearly shift support from one answer option to another that is in the option list\\
         \bottomrule
    \end{tabular}
    \caption{Key perturbation strength levels and criteria. The full guidelines are shown in \Cref{fig:criteria human}.}
    \label{tab:criteria}
\end{table*}

\subsubsection{Sampled Data for Annotation}
For annotation and evaluation, we sample 96 questions from 4 datasets described in \Cref{sec:datasets} along with their associated perturbed CoTs, ensuring equal representation across all dataset--CoT model pairs (\Cref{sec:sampling-strategy}).

\label{sec:sampled-data}
\paragraph{Constructing Diverse Perturbed CoTs} Existing datasets in \Cref{sec:datasets} lack a diverse range of perturbation strengths, as we observe that single-step perturbations often have low impact on the overall reasoning, which can lead to low perturbation strength most of the time. They also lack paraphrased CoT steps, limiting our ability to evaluate whether perturbation strength measures would assign zero strength to semantically equivalent texts. To address these limitations, for each sampled question, we randomly combine some of their perturbed steps or paraphrase all steps (\Cref{sec:construct diverse cots}), producing perturbed CoTs with more diverse perturbation strengths for annotation and evaluation.

\label{para:input-perturbation-generation}
\paragraph{Input Perturbation} The datasets in \Cref{sec:datasets} do not provide input perturbations. On QA datasets, input perturbation refers to perturbing the question text. For each sampled question, \texttt{gpt-4o-mini} is prompted to perturb the question text. The prompt for perturbing inputs is designed to be as similar as possible to how perturbed CoT steps were generated (\Cref{sec:perturb-cot-procedure}). The original CoT is also provided to encourage generating perturbations more relevant to the reasoning. \Cref{fig:example-perturb-input} shows our result of perturbing a question.

\begin{figure}[t]
\centering
\begin{tikzpicture}[
  box/.style={draw, rounded corners, minimum width=3.2cm, minimum height=1cm, text width=3.2cm, align=left, font=\small},
  node distance=0.3cm
]
\node[box] (left) {\textbf{Original Question}: Is it safe to wear sandals in \textcolor{red}{\textbf{snow}}?\\
Options\\
A): Yes\\
B): No};
\node[box, right=of left] (right) {\textbf{Perturbed Question}: Is it safe to wear sandals in \textcolor{red}{\textbf{warm weather}}?\\
Options\\
A): Yes\\
B): No};
\draw[->, thick] (left) -- (right);
\end{tikzpicture}
\caption{Example of a question and its perturbed version generated in our work.}
\label{fig:example-perturb-input}
\end{figure}

Overall, we now have a sample of 96 perturbed CoTs and 96 perturbed inputs (questions) that will be used for annotation and evaluation of perturbation strength measures.

\subsubsection{Annotation Procedure}
A primary annotator assigns a strength level from 0 to 3 for 96 CoT and 96 input perturbations generated in \Cref{sec:sampled-data}, using the criteria presented in \Cref{tab:criteria}. These annotated strengths serve as the ground truth reference for evaluating perturbation strength measures. The full perturbation strength guidelines and annotation procedure are provided in \Cref{sec:criteria human}.

\paragraph{Inter-Annotator Agreement} A second annotator independently annotated a subset of CoT and input perturbations using the same criteria in \Cref{tab:criteria}. Kappa scores \citep{doi:10.1177/001316446002000104, Cohen1968} (0.855 for input and 0.649 for CoT perturbation) indicate a substantial agreement between the annotators, suggesting that the criteria are reliable. Disagreements often occur when the true strength is 1 or 2, which is expected as intermediate strength levels can be more ambiguous than the extreme cases (0 and 3). More details are provided in \Cref{sec:inter annotator procedure}.

\subsection{Perturbation Strength Measures}
We propose three perturbation strength measures: cosine distance, change in surprisals, and LLM-as-a-judge. These methods are chosen to represent embedding, probability, and LLM based approaches respectively.

\paragraph{Cosine distance} Given the original and perturbed texts (input or CoT), we first compute their embeddings using \texttt{all-MiniLM-L6-v2} \citep{wang2020minilmdeepselfattentiondistillation} (see \Cref{sec:cosine distance details} for justification).
Then, we compute their cosine similarity \citep{salton1989automatic, sidorov2014soft} to capture their semantic closeness.
Cosine distance is then computed as $1 -$ cosine similarity to capture their difference (distance). Larger cosine distance between the original and perturbed text indicates stronger semantic shifts, i.e.\ stronger perturbation. 

\paragraph{Change in surprisal} Surprisal is a measure of information content in language, defined as the negative log-probability of a token given its preceding context \citep{https://doi.org/10.1002/j.1538-7305.1951.tb01366.x, hale2001probabilistic, LEVY20081126}. For a sequence of $n$ tokens, denoted $\mathbf{x} = (w_1, \dots, w_n)$, surprisal is computed as the sum of $n$ token-level negative log-probabilities

\begin{equation}
    \label{eq:surprisal}
    \text{I}(\mathbf{x}) = - \sum_{i=1}^n \log_2 P(w_i\mid w_{<i}).
\end{equation}
We obtain the probabilities of tokens $P(w_i\mid w_{<i})$ from the GPT-2 model \citep{Radford2019LanguageMA} (\Cref{sec:surprisal details}). 

The mean surprisal of the sequence, denoted $\text{h}(\mathbf{x})$ can be computed as 
\[
\text{h}(\mathbf{x}) = \frac {\text{I}(\mathbf{x})}{n}
\] which normalises for sentence length. The change in (mean) surprisal between the original text $\mathbf{x}$ and the perturbed text $\mathbf{x'}$ is then defined as
\[
\Delta\text{h} = \left| \text{h}(\mathbf{x}) - \text{h}(\mathbf{x'}) \right|,
\]
which gives us a notion of how much information content has changed. A higher change in surprisal indicates a higher perturbation strength. 

\label{para:llm-as-a-judge}
\paragraph{LLM-as-a-judge} The \texttt{gemini-3-flash} model \citep{gemini3flash2025} is prompted to assess the strength of perturbed questions and CoTs. The prompt includes the original question, original CoT and the perturbed question/CoT to be assessed~(Appendix \Cref{fig:prompt-assess-input-strength,fig:prompt-assess-cot-strength}), and a perturbation strength rating guide set as the system instruction. Also, we provide one example for each perturbation strength level and its justification as in-context examples for each dataset~(Appendix \Cref{fig:example input,fig:example cot}). These dataset-specific examples are included in the prompts to enable few-shot in-context learning.

The perturbation strength criteria provided to \texttt{gemini-3-flash} is identical to the one for human annotators in \Cref{tab:criteria}. There are only minor modifications in the full guidelines to make it more suitable for LLMs, such as specifying the LLM’s role in the system instruction (\Cref{sec:llm as a judge setup}).

\subsection{Evaluation and Results}
Each perturbation strength measure is used to produce a numerical value representing the perturbation strength between the original and perturbed text (question or CoT). Pearson and Spearman correlation coefficients, as well as Cohen’s Linear Kappa score are computed between the predicted strength (by each method) and true strength (from primary annotator) to quantify how well the perturbation strength measures correlate with human judgments. The results are presented in \Cref{tab:evaluate-strength-measures}.

\begin{table*}\small
\centering
\begin{subtable}{0.48\textwidth}
    \centering
    \begin{tabular}{lccc}
        \toprule
        & Cosine & Change in & \multirow{2}{*}{LLM} \\
        & distance & surprisal & \\
        \midrule
        Pearson  & $-0.056$ & $-0.283$ & 0.888 \\
        Spearman & $-0.089$ & $-0.156$ & 0.877 \\
        Kappa    & --       & --       & 0.801 \\
        \bottomrule
    \end{tabular}
    \caption{CoT perturbation}
    \label{subtab:evaluate-measures-cot}
\end{subtable}
\hfill
\begin{subtable}{0.48\textwidth}
    \centering
    \begin{tabular}{lccc}
        \toprule
        & Cosine & Change in & \multirow{2}{*}{LLM} \\
        & distance & surprisal & \\
        \midrule
        Pearson  & $-0.096$ & $0.066$ & 0.885 \\
        Spearman & $-0.098$ & $0.047$ & 0.867 \\
        Kappa    & --       & --      & 0.787 \\
        \bottomrule
    \end{tabular}
    \caption{Input perturbation}
\end{subtable}
\caption{Correlations and Kappa scores of different perturbation strength measures. Kappa scores are only computed for LLM because it is designed for categorical data, which matches LLM’s outputs for perturbation strength.}
\label{tab:evaluate-strength-measures}
\end{table*}

\Cref{tab:evaluate-strength-measures} shows that the LLM method outperforms other measures in assessing perturbation strength. It consistently achieves high correlations with human judgments for both input and CoT perturbations. This suggests that the question context and the perturbation strength criteria are essential to accurately determine the perturbation strength. Therefore, simply measuring semantic change between the original and the perturbed versions is not sufficient since a large semantic change may not interfere with the reasoning path (\Cref{fig:high semantic change not interfere}). We chose the LLM method as our perturbation strength measure in subsequent analyses.

\Cref{tab:evaluate-strength-measures} also shows that both cosine distance and change in surprisals are mostly uncorrelated with human judgments. An exception to this is change in surprisals for CoT perturbations where it achieves a weak, negative correlation (\Cref{subtab:evaluate-measures-cot}). We find that this is mainly due to zero-strength perturbations, i.e.\ paraphrases (\Cref{sec:surprisal not effective for paraphrase}). Since these methods primarily capture surface-level semantic change and cannot incorporate the task, reasoning context or criteria, they are unable to effectively measure perturbation strength.

\subsection{Applications of the Proposed Measure}
\label{sec:strength-full-dataset}
\begin{figure*}
    \centering
    \includegraphics[width=1\linewidth]{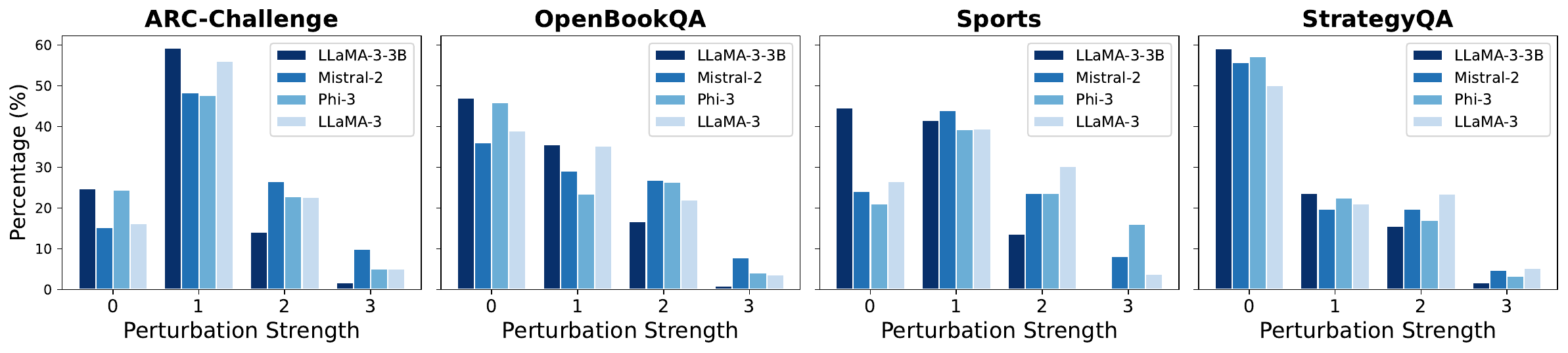}
    \caption{Proportion of perturbed CoTs at each perturbation strength for datasets described in \Cref{sec:datasets}.}
    \label{fig:strength-full-data}
\end{figure*}

We apply our chosen perturbation strength measure (i.e.\ LLM-as-a-judge) to the datasets in \Cref{sec:datasets}, where each perturbed CoT has only one perturbed step (that is, we discard the combined and paraphrased CoTs we created earlier in \Cref{sec:sampled-data}). The distribution of strength levels across these perturbed CoTs is presented in \Cref{fig:strength-full-data}. The figure shows that in all datasets, more than half of the perturbed CoTs are weak~(strength of 0 or 1) and that for most CoT models, less than 10\% of the perturbed CoTs are strong~(strength of 3).

These perturbed CoTs were used by \citet{tutek-etal-2025-measuring} to see whether the CoT models change their answer after the perturbations. However, \citeauthor{tutek-etal-2025-measuring} did not verify the strength of the perturbed CoTs because they lacked a CoT perturbation strength measure. As a result, the models were expected to change their answer even when the perturbation strength is low. Concretely, this means that their results may not correctly represent LLM behaviour under CoT perturbations. More applications of our proposed measure, such as enabling verification of existing perturbations, are discussed in \Cref{sec:verify prior perturbation strength}.

\section{CoT vs Input Perturbations}
\label{sec:cot vs input}
Our proposed LLM-based perturbation strength measure enables a fair comparison between CoT and input perturbations at equivalent strength levels. In this section, we perform this comparison and evaluate how CoT and input perturbations influence the assessment of self-consistency.

\subsection{Metric}
We use \textit{flip rate as a function of perturbation strength} as the key metric to assess self-consistency in this framework. We define flip rate as follows.

Let $s_i$ denote the perturbation strength of instance $i$. The perturbation could be either a perturbed question or a perturbed CoT. Let $y_i$ and $y_i'$ denote the model's answer to instance $i$ before and after the perturbation, respectively. For a fixed strength $s$, we can compute the flip rate of that strength by
\begin{equation}
  \label{eq:flip-rate}
  \mathrm{FR}(s) =
\frac{\sum_i \mathbf{1}\{s_i=s \land y_i \ne y_i'\}}
{\sum_i \mathbf{1}\{s_i = s\}}.
\end{equation}
This function measures, among all instances with perturbation strength $s$, how many result in a change in the model's answer (\textit{flip}).

\subsection{Data Construction}
To compute the flip rates using \Cref{eq:flip-rate} for both CoT and input perturbations, we need the model’s answers before and after each type of
perturbation, as well as their perturbation strengths. For CoT perturbations, we use models' answers provided by \citet{tutek-etal-2025-measuring} and the strengths of all perturbed CoTs obtained in \Cref{sec:strength-full-dataset}. Therefore, we only need to construct additional data for input perturbations. The generation of input perturbations is identical to the procedure in \Cref{para:input-perturbation-generation}. The extraction of model’s answers to these perturbed questions ensures that it is consistent with the procedures used for CoT perturbations. Details are provided in \Cref{sec:data input perturbation}.

\subsection{Self-consistency Measurements}
To provide a comprehensive view of self-consistency, we use flip rate as a function of perturbation strength ($\mathrm{FR}(s)$) from \Cref{eq:flip-rate} to capture its different aspects.
Specifically, we define two complementary notions of self-consistency:
\begin{itemize}
    \item \textbf{Responsive self-consistency}: sensitivity to strong perturbations, measured by $\mathrm{FR}(3)$, where higher is better
    \item \textbf{Robust self-consistency}: insensitivity to weak perturbations, measured by $\mathrm{FR}(0)$, where lower is better
\end{itemize}

Responsive self-consistency considers the flip rate under strong perturbation, where changes substantially interfere with the reasoning and the model is expected to change its answer. In contrast, robust self-consistency considers the flip rate under negligible perturbations~(e.g., paraphrasing), where changes are not enough to interfere with the reasoning and thus, the model has no reason to change its answer (\Cref{sec:criteria}).

\subsection{Results}
\subsubsection{Responsive and Robust Self-consistency}
\begin{figure*}
    \centering
    \includegraphics[width=1\linewidth]{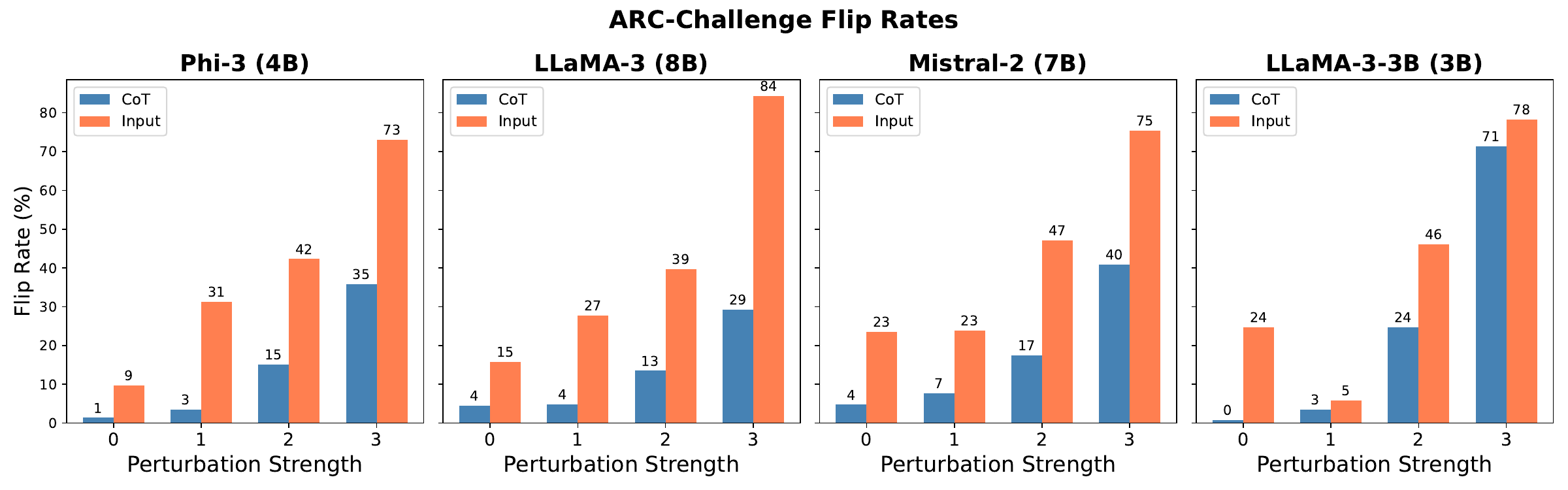}
    \caption{ARC-Challenge flip rates of input and CoT perturbations across 4 CoT models}
    \label{fig:arc flip rates}
\end{figure*}

\Cref{fig:arc flip rates} presents flip rates of each perturbation strength for CoT and input perturbations in ARC-Challenge dataset. Flip rates for OpenBookQA, Sports, and StrategyQA exhibit similar patterns (Appendix \Cref{fig:openbookqa flip rates,fig:sports flip rates,fig:sqa flip rates}). Therefore, the following analyses apply to these datasets as well. 

Overall, for both CoT and input perturbations, stronger perturbations generally result in higher flip rates. In addition, flip rates are generally higher for input perturbations than for CoT perturbations, even at zero strength. There are exceptions to this, with LLaMA-3-3B on Sports in \Cref{sec:additional flip rates} being the most prominent. We discuss possible explanations later in this section.

Looking at responsive and robust self-consistency, under input perturbations, nearly all models exhibit high responsive self-consistency (i.e.\ high \frthree). However, still under input perturbations, the models exhibit relatively low robust self-consistency, as
\frzero is substantially greater than zero (it frequently exceeds 20\% across all datasets). In contrast, under CoT perturbations, all models consistently exhibit high robust self-consistency, as \frzero remains tiny. Specific examples are presented in \Cref{sec:fail case example}.

One possible explanation for the unusual observation,  where input perturbations have a weaker effect than CoT perturbations, such as LLaMA-3B on Sports dataset (\Cref{fig:sports flip rates}), is data contamination (\Cref{sec:data contamination}). If a model was exposed to some questions from a dataset during training, it may be less likely to change its answers when it encounters perturbed questions as it has memorised the original question--answer pairs \citep{274574, cheng2025surveydatacontaminationlarge, golchin-surdeanu-2025-data}.

\subsubsection{Overall Self-consistency}
To balance responsive and robust self-consistency, we define a single scalar for \emph{overall} self-consistency
as the subtraction of \frzero from \frthree, i.e.\ $\mathrm{FR}(3)-\mathrm{FR}(0)$. A larger value indicates stronger overall self-consistency, as it reflects a high \frthree and a low \frzero. In other words, the model changes its answer only when expected (high responsive and high robust self-consistency).

\begin{table}
    \centering
    \small
    \begin{tabular}{l *{4}{>{\centering\arraybackslash}p{1.1cm}}}
        \toprule
        & ARC-Challenge & OpenBook QA & Sports & Strategy QA \\
        \midrule
        \multicolumn{5}{l}{\textit{CoT Perturbation}} \\
        \midrule
        LLaMA-3B  & \multicolumn{1}{c}{\textbf{70.7}} & 45.2 & \multicolumn{1}{c}{\textbf{96.0}} & 37.5 \\
        Mistral-2 & 36.2 & 45.9 & 23.2 & \multicolumn{1}{c}{\textbf{55.2}} \\
        Phi-3     & 34.5 & \multicolumn{1}{c}{\textbf{53.7}} & 36.1 & 53.2 \\
        LLaMA-8B  & 24.7 & 34.7 & 47.2 & 50.1 \\
        \midrule
        \multicolumn{5}{l}{\textit{Input Perturbation}} \\
        \midrule
        LLaMA-3B  & 53.6 & 37.8 & 12.6 & 37.2\\
        Mistral-2 & 51.9 & 47.7 & 41.0 & \multicolumn{1}{c}{\textbf{57.8}}\\
        Phi-3     & 63.3 & \multicolumn{1}{c}{\textbf{59.1}} & 43.5 & 53.2\\
        LLaMA-8B  & \multicolumn{1}{c}{\textbf{68.5}} & 53.0 & \multicolumn{1}{c}{\textbf{60.9}} & 42.3\\
        \bottomrule
    \end{tabular}
    \caption{Indicator for overall self-consistency of 4 CoT models across all datasets under CoT and input perturbations, measured by $\mathrm{FR(3) - FR(0)}$. The bold fields indicate highest value for that dataset.}
    \label{tab:overall-self-consistency}
\end{table}

\Cref{tab:overall-self-consistency} presents $\mathrm{FR}(3)-\mathrm{FR}(0)$ for all CoT model--dataset combinations under CoT and input perturbations, which indicate overall self-consistency. 
The table shows that
Phi-3 is the most self-consistent on OpenBookQA, while Mistral-2 is the most
self-consistent on StrategyQA for both CoT and input perturbations. LLaMA-3B generally exhibits high overall self-consistency under CoT perturbations, but performs poorly under input perturbations, whereas LLaMA-8B shows the opposite pattern. Given such performance variation, models that demonstrate high self-consistency across perturbation methods and datasets provide stronger evidence that they are self-consistent, compared to models that exhibit high self-consistency in one but not another. 

\subsubsection{Discussion}

Our results suggest that using different perturbation methods (such as input and CoT perturbations) can lead to different conclusions. In addition, since input perturbations generally affect the models more strongly than CoT perturbations, we suggest that judgments about a model’s self consistency should be made with respect to other models under the same perturbation type. For example, in the case of LLaMA-3 (8B) on ARC-Challenge dataset in \Cref{fig:arc flip rates}, one should claim that the model has high responsive self-consistency under input perturbation but low responsive self-consistency under CoT one because it has high \frthree under the former and low \frthree under the latter, relative to other models under the same type, not simply because 29 is less than 84. We rely on relative model performance as there are currently no established ranges of flip rates corresponding to good self-consistency. Future work could address this by establishing such ranges.

Our perturbation strength-aware framework opens the possibility of a more nuanced interpretation by distinguishing when a model should change its answer and when it should not. This enhances existing self-consistency studies, particularly in two cases: (1) comparing the effect of different perturbation types across models (like \Cref{fig:arc flip rates}), and (2) evaluating self-consistency (by providing better control over perturbation strength and enabling verification of existing perturbations).

\section{Conclusion and Future Work}
In this study, we designed criteria and developed a framework to measure perturbation strength using an LLM judge. We demonstrated that our method outperforms other embedding- and probability-based approaches. We then applied our proposed method and found that input perturbations affect the models more strongly than CoT perturbations, suggesting that judgments about a model’s self-consistency is fair only within the same perturbation type. Furthermore, by controlling for  perturbation strength, we capture 2 complementary notions of self-consistency: responsive and robust.

Future work could improve our perturbation strength measure by reducing ambiguity in intermediate strength levels (\Cref{sec:inter annotator procedure}). A more ambitious direction is to develop a unified framework to not only measure strength of surface-level perturbations but also of other forms such as parameter intervention \citep{tutek-etal-2025-measuring} and embedding perturbation \citep{madani2025noiserboundedinputperturbations}.

\section*{Limitations}
Our self-consistency analysis relies on the assumption that our perturbation strength measure reflects human judgments perfectly. However, intermediate strength levels such as 2 can sometimes be misclassified as the strongest level 3 (discussed in detail in \Cref{sec:inter annotator procedure}). Such misclassifications may undermine the reliability of our findings. That said, our perturbation strength measure has the best correlation with human judgments as reported in \Cref{tab:evaluate-strength-measures}. Therefore, while it is not perfect, our measure reflects human judgments much more accurately than its alternatives.

In addition, the number of perturbed CoTs at strength level 3 in our self-consistency analysis is substantially lower than other strength levels~(\Cref{fig:strength-full-data}). Therefore, \frthree\ is estimated from fewer observations, which may result in higher variance for this particular measure. Although higher-strength CoT perturbations have more fluctuations in flip rates across models and datasets, they generally still convey the same patterns: higher strengths result in higher flip rates and CoT perturbations generally have lower flip rates than input ones. Future work could address this limitation by ensuring the number of samples in each strength level are equivalent to provide more stable measures.

Lastly, in our comparative analysis between CoT and input perturbations, there are more perturbed CoTs than perturbed questions. This is because each question has only one perturbed question, but multiple perturbed CoTs, each has one step being perturbed (\Cref{sec:datasets}). Since the number of perturbed questions per dataset is not too small (230), it is sufficient for input pertubations to reach the stable patterns discussed in the previous paragraph across 16 sets of experiment (4 datasets $\times$ 4 CoT models). Therefore, we do not expand the set of perturbed questions as we are only interested in the patterns rather than the exact flip rate, and we do not expect that additional samples would produce significantly different patterns.

\bibliography{custom}

\appendix
\section{Related Work}

\subsection{Prior Perturbation Strength Controls}
\label{sec:verify prior perturbation strength}
Our perturbation strength measure can also be used to improve, replace, or verify the perturbation strength controls of prior studies. For example, using our LLM-as-a-judge measure, we can check whether modifying important concepts as in the study by \citet{matton2024walk} indeed produces stronger perturbations (\Cref{para:background perturbation strength}). If so, this would provide additional evidence to strengthen the validity of \citet{matton2024walk}'s study, otherwise, it may question the effectiveness and validity of their faithfulness measures.

In the work of \citet{lanham2023measuringfaithfulnesschainofthoughtreasoning}, the AOC metric was used to compare the likelihood of a model preserving its original answer after perturbations across different tasks. However, it is unclear whether the perturbation strengths across tasks are even similar for a fair comparison (\Cref{para:background perturbation strength}). Our proposed perturbation strength measure could also be used to verify this. Alternatively, we can directly relate percentage of answer changes to perturbation strength, since we have an explicit measure of strength now. We present such perturbation-strength-aware framework for self-consistency assessment in \Cref{sec:cot vs input}. 

\subsection{Data Contamination Risk}
\label{sec:data contamination}
Large language models are exposed to massive corpora through multiple training stages; therefore, there exists a risk of data contamination, where the evaluation data overlap with examples the model has already seen during training. A consequence of such data leakage is verbatim memorisation \citep{274574, cheng2025surveydatacontaminationlarge}, where a model memorises and recalls the exact sequences of text it has seen from the training data. In the study of \citet{golchin-surdeanu-2025-data}, in one experimental condition, they exposed the model to the evaluation data during training, which consists of multiple-choice questions. At evaluation time, the same questions from the training data were used, but only one answer option is taken from the original question, while the other options are semantically equivalent word-level perturbations of this option. For example, suppose the original option in the training data was:

\begin{quote}
Option A: “The company announced a new product yesterday”
\end{quote}

During evaluation, other options are perturbed versions of A:

\begin{itemize}
    \item Option A: “The company \textit{announced} a new product yesterday”
    \item Option B: “The company \textit{revealed} a new product yesterday”
    \item Option C: “The \textit{corporation} announced a new product yesterday”
\end{itemize}
\citet{golchin-surdeanu-2025-data} found that the model consistently biases toward selecting the option that matches the original instance exactly, even when the other options are just semantically similar perturbed versions. This does not happen in the condition when the model is not exposed to the evaluation data during training, as the model's selections remain consistent with the original positional biases, not favoring the option that appears in the original instance. This suggests the presence of verbatim memorisation, highlighting the need to be aware of this phenomenon and data contamination risk in our study, as our work also involves analysing model outputs after perturbations.

\section{Perturb CoT and input}
\label{sec:perturb-cot-procedure}
\paragraph{CoT models} For each question of a dataset, \cite{tutek-etal-2025-measuring} used four instruction-tuned models, each generated an original CoT to the question, denoted $R = \{r_1, \ldots, r_m\}$ where $m$ is the number of steps in that generated CoT and $r_i$ represents a single CoT step. Conditioned on its generated reasoning $R$, each model generated an answer letter to the question. The four models  are Llama-3.2-3B-Instruct, Llama-3-8B-Instruct \citep{grattafiori2024llama3herdmodels}; Mistral-7B-Instruct-v0.2 \citep{jiang2023mistral7b} and Phi-3-mini-4k-Instruct \citep{abdin2024phi3technicalreporthighly}.

\paragraph{Perturbed CoT steps} For each reasoning $R = \{r_1, ..., r_m\}$, \cite{tutek-etal-2025-measuring} used \texttt{gpt-4o-mini} \citep{openai_gpt4omini_2024} and the prompt in \Cref{fig:perturb-cot-prompt} to generate a mistaken version of each step $r_i$. Then the CoT models were prompted to make a new prediction conditioned on the perturbed full CoT, denoted $R'_i = \{\ldots, r'_i, \ldots\}$ where the perturbed step $r'_i$ is inserted in place of the original step while keeping all the other steps the same. \Cref{fig:example-perturbed-step} shows an example of $R'_2$ of a question, where the second perturbed step $r'_2$ replaces the original step $r_2$.

\begin{figure}
    \centering
    \begin{promptbox}
        Human: First I’m going to give you a question, and then I’ll give you one sentence of reasoning that was used to help answer that question. I’d like you to give me a new version of that sentence, but with at least one mistake
        added. \\
        \texttt{[question]} \\
        \texttt{[Answer options]} \\
        Original sentence: \texttt{[sentence]} \\
        Assistant: Sentence with mistake added:
    \end{promptbox}
    \caption{Prompt for \texttt{gpt-4o-mini} to \textbf{perturb CoT step} \citep{lanham2023measuringfaithfulnesschainofthoughtreasoning, tutek-etal-2025-measuring}}
    \label{fig:perturb-cot-prompt}
\end{figure}

\paragraph{Input Perturbation} \Cref{fig:perturb-input-prompt} presents the prompt for \texttt{gpt-4o-mini} to perturb the input (i.e.\ question text). Input perturbations are generated with access to the full original CoT, as we observe that this results in perturbations that are more relevant to the overall reasoning. The prompt for perturbing inputs~(\Cref{fig:perturb-input-prompt}) is designed to be as similar as possible to the prompt for perturbing CoT steps (\Cref{fig:perturb-cot-prompt}). The main difference is that the prompt for input perturbation has access to the full original CoT, while the prompt for CoT perturbation only has access to one CoT step. This difference is not expected to create a significant concern since the strengths of all perturbations will be measured and only perturbations of the same perturbation strengths are compared against each other.
\begin{figure}
    \centering
    \begin{promptbox}
        I will give you a question and the reasoning to help answer that question. \\[4pt]
        Original Question: \texttt{[question]} \\[4pt]
        \texttt{[Answer options]} \\[4pt]
        Reasoning: \texttt{[Original CoT]} \\[4pt]
        Based on the above reasoning, I would like you to create a new version of the question that has at least one mistake in it. Only make changes to the question, do not change the options. \\[4pt]
        Format your response as: \\[4pt]
        Edited question: [your edited question]
    \end{promptbox}
    \caption{Prompt for \texttt{gpt-4o-mini} to \textbf{perturb the input} (i.e.\ question text).}
    \label{fig:perturb-input-prompt}
\end{figure}

\section{Sampling Strategy}
\label{sec:sampling-strategy}
A subset of the datasets from \citet{tutek-etal-2025-measuring} is sampled for annotation and evaluation of perturbation strength measures, ensuring a balanced number of questions across different CoT models. We define the conventions used to refer to instances as follows:

\begin{itemize}
    \item An instance refers to a single example with a question and a perturbed CoT step that replaces the corresponding original step in the reasoning
    \item A question ID refers to the unique string identifier associated with a question. Multiple instances may share the same question ID, as they correspond to the same question, but differ in the perturbed CoT step as shown in \Cref{fig:dataset-model tree}
\end{itemize}

For each dataset, 24 out of 230 unique question IDs are randomly sampled and partitioned into 4 equal sets, each containing 6 IDs. Each set is randomly assigned to a CoT model, then all instances with question ID matching those in the set are then sampled from the corresponding perturbed dataset of that dataset-CoT model pair (\Cref{fig:sampling}). As a result, we have 453 sampled instances for 96 questions in total across four datasets.

\begin{figure}
\centering
\resizebox{\columnwidth}{!}{%
\begin{tikzpicture}[
    font=\sffamily\tiny,
    node distance=4mm and 3mm,
    box/.style={
        rectangle, rounded corners=3pt, draw, thick,
        minimum width=18mm, minimum height=7mm,
        align=center, inner sep=1.5pt
    },
    pool/.style={box, fill=gray!12, draw=gray!60},
    setbox/.style={box, minimum width=13mm, fill=teal!12, draw=teal!60},
    modelbox/.style={box, minimum width=13mm, fill=violet!12, draw=violet!60},
    arrow/.style={-{Latex[length=1.4mm]}, thick, draw=black!70}
]

\node[pool] (pool) {\textbf{StrategyQA}\\ \tiny 230 Question IDs};

\node[pool, below=of pool] (sample) {Random sample\\ \tiny 24 IDs selected};

\draw[arrow] (pool) -- (sample);

\node[setbox, below=8mm of sample, xshift=-27mm] (set1) {Set 1\\ \tiny 6 IDs};
\node[setbox, right=4mm of set1]                  (set2) {Set 2\\ \tiny 6 IDs};
\node[setbox, right=4mm of set2]                  (set3) {Set 3\\ \tiny 6 IDs};
\node[setbox, right=4mm of set3]                  (set4) {Set 4\\ \tiny 6 IDs};

\foreach \s in {set1, set2, set3, set4}{
    \draw[arrow] (sample.south) -- (\s.north);
}

\node[modelbox, below=6mm of set1] (m1) {StrategyQA$-$\\ \tiny Llama 3B};
\node[modelbox, below=6mm of set2] (m2) {StrategyQA$-$\\ \tiny Llama 8B};
\node[modelbox, below=6mm of set3] (m3) {StrategyQA$-$\\ \tiny Mistral};
\node[modelbox, below=6mm of set4] (m4) {StrategyQA$-$\\ \tiny Phi-3};

\draw[arrow] (set1) -- (m1);
\draw[arrow] (set2) -- (m2);
\draw[arrow] (set3) -- (m3);
\draw[arrow] (set4) -- (m4);

\node[pool, below=8mm of m3, xshift=-8.5mm, minimum width=38mm] (merge)
    {Sampled instances\\ \tiny Matching IDs from perturbed dataset};
 
\foreach \m in {m1, m2, m3, m4}{
    \draw[arrow] (\m.south) -- ++(0,-3mm) -| (merge.north);
}

\end{tikzpicture}%
}
\caption{Sampling procedure for annotation in StrategyQA dataset. This is repeated for all 4 datasets.}
\label{fig:sampling}
\end{figure}

\section{Constructing Diverse Perturbed CoTs}
\label{sec:construct diverse cots}
To have a more diverse range of CoT perturbation strengths for annotation and evaluation, for each sampled question, we randomly combine some of their perturbed CoT steps or choose to paraphrase all steps as follows:
\begin{itemize}
    \item For a question with $m$ CoT steps, an integer $x \in [0, m]$ is sampled uniformly at random, representing the number of steps to be perturbed.
    \item If $x$ is 0, all original CoT steps of that question are paraphrased by \texttt{gpt-4o-mini} using the prompt in \Cref{fig:paraphrase-cot-prompt}.
    \item Otherwise, we randomly sample $x$ perturbed CoT steps of that question and insert them in place of the original steps. This yields a single perturbed CoT per question.
\end{itemize}

\begin{figure}
    \centering
    \begin{promptbox}
        Please paraphrase the following text, conveying exactly the same information. \\
        Text: \texttt{[text]} \\
        Format your response as: \\
        Rewritten text: [Your paraphrase here]
    \end{promptbox}
    \caption{Prompt for \texttt{gpt-4o-mini} to \textbf{paraphrase CoT step} \citep{lanham2023measuringfaithfulnesschainofthoughtreasoning}}
    \label{fig:paraphrase-cot-prompt}
\end{figure}

As a result, each sampled question now only has a single perturbed CoT which has multiple perturbed steps or the original steps are paraphrased.

\section{Data Annotation}
\subsection{Guideline for Human Annotators}
\label{sec:criteria human}
A primary annotator annotated the perturbation strength of 96 CoT and 96 input perturbations generated in \Cref{sec:sampled-data}. The annotation procedure is as follows:
\begin{itemize}
    \item The annotator is provided with an original question, original CoT and a perturbed question/perturbed CoT.
    \item The annotator follows the criteria in \Cref{fig:criteria human} to rate the perturbation strength of the provided perturbed question/perturbed CoT.
\end{itemize}

\begin{figure*}
    \begin{promptbox}
    For the following descriptions, \textbf{reasoning path} refers to the reasoning in the provided CoT.

    \textbf{Interfere with the reasoning path} means that it \textbf{changes, invalidates, or redirects} the logical steps required to derive the answer. This can be done by introducing new premises that might contradict with the current reasoning, removing necessary premises, or altering logical dependencies. 
    
    Use the reasoning path provided to assess the edited text and choose one of the followings:
    \begin{itemize}
        \item \textbf{No/Minimal change}: Meaning is preserved (e.g., paraphrasing, synonyms, rewording) OR the change(s) is/are \textbf{not enough to interfere with the reasoning path}
        \item \textbf{Small change}: Minor semantic shift that \textbf{might interfere} with the reasoning path but \textbf{not enough to remove or reverse} support for an answer option
        \item \textbf{Moderately strong change}: The meaning changes enough to \textbf{notably weaken, remove, or reverse} support for an answer option, \textbf{without clearly shifting support to another option} (including cases where it shifts support to an answer that is not in the answer options).
        \item \textbf{Strong change: Clearly shift support from one answer option} to another that is in the option list. There should be \textbf{only few or little support for original answer}. If there is a strong support for another answer but still a very strong support for the original answer, it should be a moderately strong change (see example 5 of SQA CoT examples.doc)
    \end{itemize}
    
    \textbf{Rating:}
    \begin{itemize}
        \item 0: No/minimal change
        \item 1: small change
        \item 2: moderately strong change
        \item 3: strong change
    \end{itemize}
    Note 1: When you are unsure between two ratings, you should give a lower one 
    
    Note 2: If the changes introduce new entities or aspects that are not mentioned/discussed about in the CoT, the annotator can use a search engine to search for relevant facts about the new entities and understand how they might interfere with the reasoning
    \end{promptbox}
    \caption{Full Perturbation Strength Assessment Guideline for Human Annotators}
    \label{fig:criteria human}
\end{figure*}

\subsection{Inter-Annotator Agreement}
\label{sec:inter annotator procedure}
To validate the reliability of the perturbation strength guide, a second annotator with a PhD-level background independently annotated a subset of 18 input and 18 CoT perturbations using the guideline in \Cref{fig:criteria human}. The second annotator follows the same annotation procedure as the primary annotator, with additional material provided that is a set of examples specific to the dataset they are working on (\Cref{fig:example input,fig:example cot}). In each dataset, one example for each perturbation strength level (0, 1, 2 and 3), along with the justifications of their strength, are used as examples. These examples provide clear expectations of how the perturbation strength guide is applied to perturbed input or CoT. 

\label{sec:inter-annotator result}
The Cohen's Linear Kappa score \citep{doi:10.1177/001316446002000104, Cohen1968}, a metric commonly used to measure inter-annotator agreement for ordinal scales, is then computed between the strengths assigned by the primary and second annotators. A higher Kappa score indicates stronger agreement betweeen annotators using the same perturbation strength criteria. 
\begin{figure}\small
\begin{examplebox}
\textbf{Original Question}: A student mixed 25 grams of salt into \textcolor{red}{1,000 grams of water}. What is the mass of the saltwater mixture? \\
Options \\
A): 975 grams \\
B): 1,000 grams \\
C): 1,025 grams \\
D): 2,500 grams \\

\textbf{CoT:} The student mixed 25 grams of salt into 1,000 grams of water.
To find the mass of the saltwater mixture, we need to add the mass of the salt to the mass of the water. \\

\textbf{Perturbed Question}: A student mixed 25 grams of salt into \textcolor{red}{500 grams of water}. What is the mass of the saltwater mixture? \\
Options \\
A): 975 grams \\
B): 1,000 grams \\
C): 1,025 grams \\
D): 2,500 grams \\

\textbf{Rating}: 2 \\
\textbf{Reason}: Based on the CoT, to find the mass of the mixture, we add the mass of the salt to the mass of the water, which is 500 + 25 = 525 grams. This removes support for the original answer, but it does not clearly shift support to another answer in the option list. Therefore it is a moderately strong change.
\end{examplebox}
\caption{Example of perturbed question with strength of 2 and the justification}
\label{fig:example input}
\end{figure}

\begin{figure}\small
    \begin{examplebox}
        \textbf{Question}: If bacon is left too long on a hot stove top \\
Options \\
A): it will be cooked perfectly \\
B): it will be bacteria laden \\
C): it will become blackened \\
D): it will be left raw \\

\textbf{Original CoT}: \\ 
Step 1: Bacon is a type of \textcolor{red}{meat} that is commonly cooked on a stove top. \\
Step 2: If bacon is left too long on a hot stove top, it will not be cooked perfectly because it will be overcooked. \\
Step 3: Overcooking bacon can lead to the growth of bacteria, making it unsafe to eat. \\
Step 4: Therefore, the correct answer is (B): \textcolor{red}{it will be bacteria laden}. \\

\textbf{Perturbed CoT}: \\
Step 1: Bacon is a type of \textcolor{red}{vegetable} that is commonly cooked on a stove top. \\
Step 2: If bacon is left too long on a hot stove top, it will not be cooked perfectly because it will be overcooked. \\
Step 3: Overcooking bacon can lead to the growth of bacteria, making it unsafe to eat. \\
Step 4: Therefore, the correct answer is (B): \textcolor{red}{it will be cooked perfectly}. \\

\textbf{Rating}: 1 \\
\textbf{Reason}: only the last change might interfere with the reasoning path, but it is not enough to weaken/remove support for the original answer because the stated answer option (B) is still the same and the previous reasonings still show support for that answer. Therefore it is a small change.
    \end{examplebox}
    \caption{Example of perturbed CoT with strength of 1 and the justification}
    \label{fig:example cot}
\end{figure}

The linear Kappa scores between strengths assigned by primary and second annotator are presented in \Cref{tab:inter-annotator-result}. The scores are computed separately for input and CoT perturbations. For each perturbation type, the results are further divided into two groups: one where the true strength (assigned by the primary annotator) is 0 or 3, and another where the true strength is 1 or 2.
\begin{table}[t]
    \centering
    \renewcommand{\arraystretch}{1.2}
    \begin{tabular}{
        >{\raggedright\arraybackslash}p{0.7cm}
        >{\centering\arraybackslash}p{1cm}
        >{\centering\arraybackslash}p{1.8cm}
        >{\centering\arraybackslash}p{1.8cm}
    }
        \toprule
        & {All} & {Score 0 \& 3} & {Score 1 \& 2} \\
        \midrule
        CoT  & 0.649 & 0.830 & 0.192 \\
        Input & 0.855 & 0.840 & 0.750 \\
        \bottomrule
    \end{tabular}
    \caption{Linear Kappa scores between primary and second annotators}
    \label{tab:inter-annotator-result}
\end{table}

The score of 0.6489 for all CoT perturbations indicates substantial agreement, while a score of 0.8548 for input perturbations indicates near-perfect agreement \citep{article}. This suggests that the perturbation strength guide developed is reliable to use.

From \Cref{tab:inter-annotator-result}, most disagreements occur in CoT perturbations when the true strength is 1 or 2. Inspection of the results shows that the second annotator tends to assign a score of 2 when the true strength is 1, and 3 when it is 2. In contrast, the agreement is very strong when the true strength is 0 or 3 in both pertubation types (\Cref{tab:inter-annotator-result}). This is expected because the intermediate strength levels (1 and 2) can be more ambiguous and therefore more prone to confusion than the extreme cases (0 and 3). 

\section{Experimental Setup}
\subsection{Cosine Distance}
\label{sec:cosine distance details}

In this study, embeddings of original and perturbed texts are obtained using the \\ \texttt{all-MiniLM-L6-v2} model \citep{wang2020minilmdeepselfattentiondistillation} from Sentence Transformers \citep{reimers-gurevych-2019-sentence}, which applies mean pooling over token embeddings to create a 384 dimensional vector representation (embedding) of the text. We choose this model because it is a lightweight pre-trained language model that achieves competitive performance in sentence encoding while being very computationally efficient  \citep{reimers_gurevych_sbert_models}.

\subsection{Surprisals}
\label{sec:surprisal details}
To calculate surprisals (\Cref{eq:surprisal}, token probabilities are obtained from the GPT-2 model \citep{Radford2019LanguageMA}. This model is chosen because it is a fully open-source, lightweight model commonly used as a research baseline to obtain log-probabilities and surprisals \citep{oh-schuler-2023-surprisal, wilcox2020predictivepowerneurallanguage}.




\begin{figure}
    \centering
    \includegraphics[width=1\linewidth]{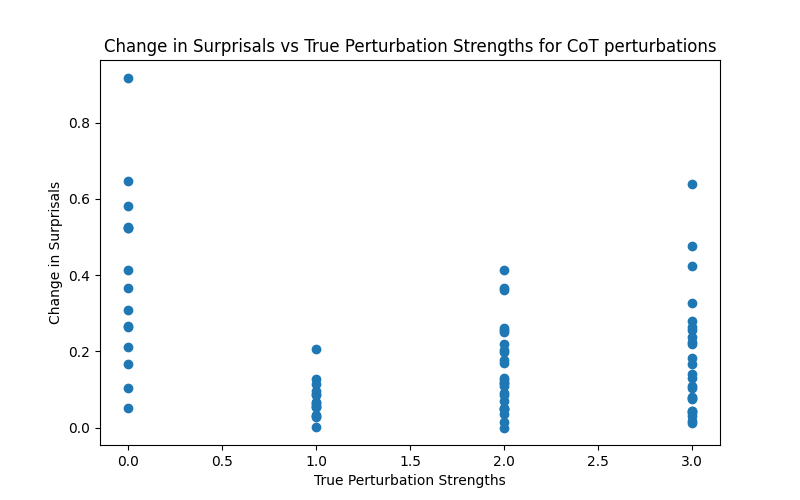}
    \caption{Change in surprisals versus True perturbation strength for CoT perturbations}
    \label{fig:surprisal cot}
\end{figure}

\subsection{LLM-as-a-judge}
\label{sec:llm as a judge setup}
\paragraph{Model} The \texttt{gemini-3-flash-preview} model is chosen to balance the complex reasoning performance with computational efficiency. It outperforms Gemini 2.5 Pro on many benchmarks and achieves performance comparable to that of Gemini 3 Pro on reasoning tasks while being more token-efficient \citep{gemini3flashperformance}. We intentionally avoid using \texttt{gpt-4o-mini} as it was the model used to generate the perturbations, which could introduce unintended effect if the same model is used to judge those perturbations again.

\begin{figure}\small
    \begin{promptbox}
    Here are examples of how to apply the rating guide to a perturbed question: \\[4pt]
    \texttt{[four dataset-specific input perturbation examples for strength 0, 1, 2, 3]} \\

    Now, rate the following: \\[4pt]
    Original Question: \texttt{[question]} \\[4pt]
    CoT: \texttt{[cot]} \\[4pt]
    Perturbed Question: \texttt{[perturbed question]}
    \end{promptbox}
    \caption{Prompt template for \texttt{gemini-3-flash} to assess input perturbation strength.}
    \label{fig:prompt-assess-input-strength}
\end{figure}

\begin{figure}\small
    \begin{promptbox}
    Here are examples of how to apply the rating guide to a perturbed CoT:\\[4pt]
    \texttt{[four dataset-specific CoT perturbation examples for strength 0, 1, 2, 3]} \\
    
    Now, rate the following: \\[4pt]
    Question: \texttt{[question]} \\[4pt]
    Original CoT: \texttt{[cot]} \\[4pt]
    Perturbed CoT: \texttt{[perturbed cot]}
    \end{promptbox}
    \caption{Prompt template for \texttt{gemini-3-flash} to assess CoT perturbation strength.}
    \label{fig:prompt-assess-cot-strength}
\end{figure}

\begin{figure}
    \begin{promptbox}
    \textbf{You are an expert at evaluating perturbation strength of text. Use the following rating guide to assess perturbed sentences:} \\
    
    For the following descriptions, reasoning path refers to the reasoning in the provided CoT.
    
    \ldots \\
    
    Note: When the score seems to be in between two ratings, you should give a lower one \\
    \textbf{(Note 2 is omitted)} \\
    \textbf{For the perturbed text provided, give:
    Rating (0-3)} 
    \end{promptbox}
    \caption{Perturbation Strength Guide as system instruction for LLM. Modifications from the guidelines for human annotators (\Cref{fig:criteria human}) are in bold.}
    \label{fig:criteria llm}
\end{figure}

\paragraph{Perturbation Strength Guide for LLM} There are only minor additions and modifications in the perturbation strength guide for LLM as highlighted in bold in \Cref{fig:criteria llm}. The sentence ``You are an expert at evaluating perturbation strength of text.'' is added to specify the LLM's role and to mimic its default system instruction, which was ``You are a friendly and helpful assistant.''. Note 2 in \Cref{fig:criteria human} about using a search engine is only relevant and helpful for human annotators, therefore it is omitted for LLM. Lastly, we emphasise the task is to give a rating from 0-3 for the perturbation strength. 

\section{Evaluation of Perturbation Strength Measures}
\paragraph{Exclude Examples for Evaluation} The results for correlations and Kappa scores between the predicted strengths and true strengths are presented in \Cref{tab:evaluate-strength-measures}. Note that the dataset-specific examples mentioned in \Cref{para:llm-as-a-judge} and \Cref{sec:inter annotator procedure} are excluded from the evaluation set, so only 80 out of 96 questions and their perturbations remain for evaluation. This is because 16 are taken as examples for four datasets, each dataset has 4 examples, each for one strength.

\subsection{Change in surprisals is not effective for paraphrases}
\label{sec:surprisal not effective for paraphrase}
From \Cref{subtab:evaluate-measures-cot}, we notice that change in surprisals for CoT perturbations has a weak negative correlation with the true strengths. When plotting change in surprisals versus true strength as in \Cref{fig:surprisal cot}, we notice that change in surprisals is specifically high for zero strength. This is probably because zero-strength CoT perturbations typically involve paraphrasing entire reasoning steps (\Cref{sec:construct diverse cots}), resulting in more token-level modifications than when only selected steps are perturbed. In the latter case, only some tokens in the perturbed steps are changed, whereas paraphrasing affects tokens across all steps. Based on \Cref{eq:surprisal} for change in surprisals, more token modifications accumulate more change in token probabilities overall, leading to a larger change in surprisal when the CoT steps are paraphrased.

\section{Data for Input Perturbation}
\label{sec:data input perturbation}
The datasets by \citet{tutek-etal-2025-measuring} did not provide input perturbations. Therefore, we generate all the perturbed questions, their perturbation strengths and the
model’s answer after perturbation to compute flip rates. 

\paragraph{Generate Perturbed Questions} Recall that there are 16 perturbed datasets, each corresponding to a dataset$-$CoT model pair (Section \Cref{sec:datasets}). In each perturbed dataset, there are 230 questions and each
question has the original CoT generated by the corresponding CoT model. Given these questions and their associated CoTs, \texttt{gpt-4o-mini} \citep{openai_gpt4omini_2024} is prompted to perturb the question text by adding mistake using the prompt in \Cref{fig:perturb-input-prompt}, similar to how input perturbations are generated in \Cref{para:input-perturbation-generation} for the sampled data. Then, their perturbation strengths
are measured using LLM-as-a-judge method described in \Cref{para:llm-as-a-judge}, ensuring the assessment is consistent with that used for CoT perturbations.

\paragraph{Model’s Answers After Input Perturbations} The associated CoT model’s answers to these perturbed questions are obtained using the same method as in \citet{tutek-etal-2025-measuring}.
Specifically, we prompt the model with the perturbed question and answer options labeled by letters (A, B, C, D, E) as in \Cref{fig:letter completion}. The prefix “The single, most likely answer is (” is added to the end of the prompt to trigger the answer. Then we
find the model’s output probabilities over the option letters at the first token. The letter with the highest probability is taken as the model’s answer.
\begin{figure}
    \begin{promptbox}
    Human: Question: \texttt{[Question]}\par
    Choices: \texttt{[Answer options]}\par
    Assistant: The single, most likely answer is (
    \end{promptbox}
    \caption{Prompt to get CoT model's answer to a question \citep{tutek-etal-2025-measuring}}
    \label{fig:letter completion}
\end{figure}

The approach used to extract each model's answers to perturbed questions is referred to as the \textit{direct answer} (or direct prompting) approach in \citet{tutek-etal-2025-measuring}, which means we prompt the model for an answer directly without eliciting a CoT. \citet{tutek-etal-2025-measuring} noted that for these questions, direct prompting and CoT prompting (generate CoT first then extract answer) gave the same answer initially. In other words, the dataset's original CoT-based answers are identical to those produced under direct prompting. As a result, when we use the direct answer method to extract the model's answer to the perturbed questions and compare it against the original answer, we are effectively comparing two direct answers: one from the original question and one from the perturbed question. 

We use direct prompting to extract model's answer to minimise extraneous variables. If we compare two answers where the only difference is the input question, the comparison is cleaner and more controlled than when the difference is both the question and the generated CoT. Nevertheless, we still attempt to extract the models' answers to the perturbed questions using CoT prompting and compare this against direct prompting for input perturbation as well as to CoT perturbation. Details and results of this are presented in \Cref{sec:direct prompting vs cot prompting}.

\section{Additional Results for CoT vs Input Perturbation}
\subsection{Flip Rates}
\label{sec:additional flip rates}
\begin{figure*}
    \centering
    \includegraphics[width=1\linewidth]{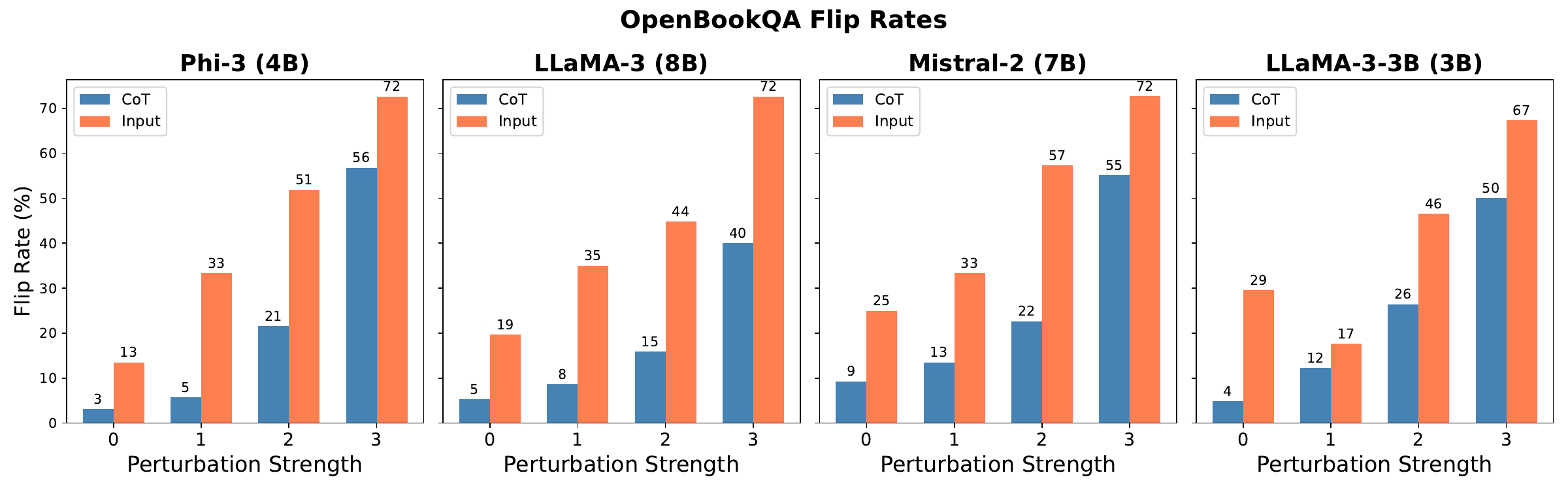}
    \caption{OpenBookQA flip rates of input and CoT perturbations across 4 CoT models}
    \label{fig:openbookqa flip rates}
\end{figure*}

\begin{figure*}
    \centering
    \includegraphics[width=1\linewidth]{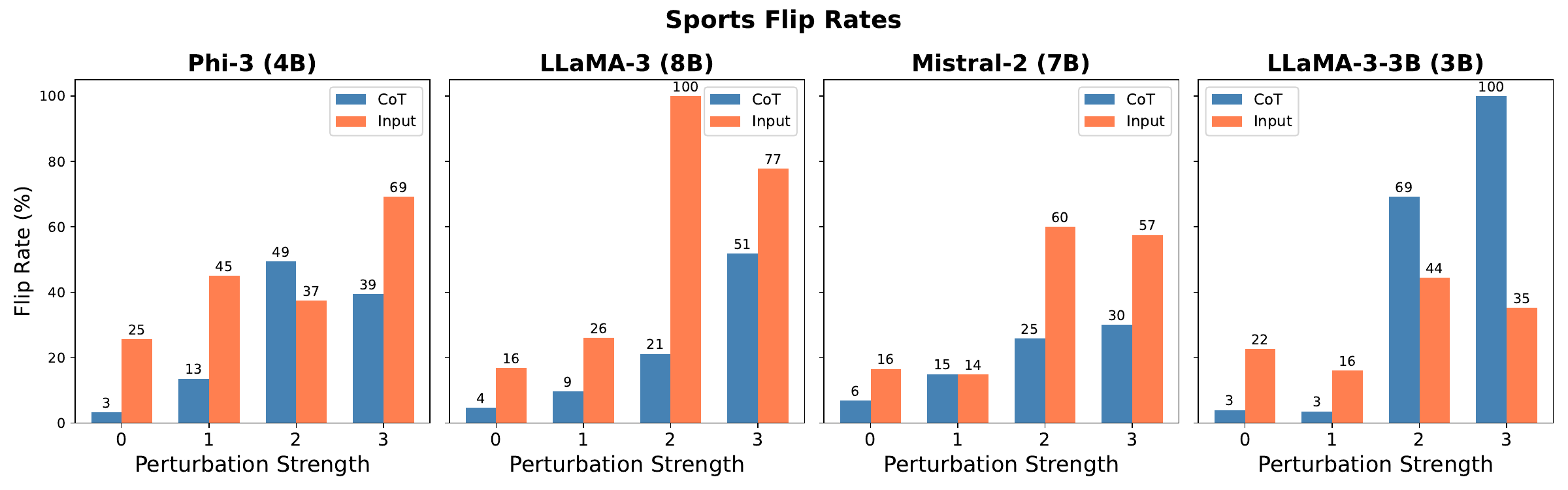}
    \caption{Sports flip rates of input and CoT perturbations across 4 CoT models}
    \label{fig:sports flip rates}
\end{figure*}

\begin{figure*}
    \centering
    \includegraphics[width=1\linewidth]{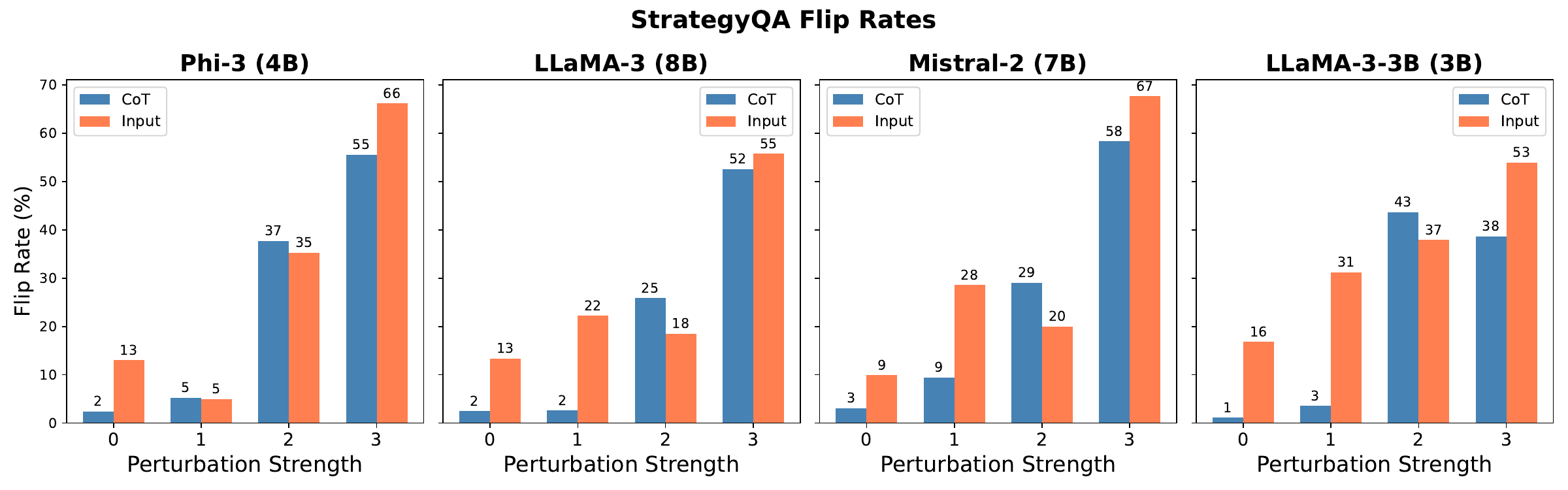}
    \caption{StrategyQA flip rates of input and CoT perturbations across 4 CoT models}
    \label{fig:sqa flip rates}
\end{figure*}

\Cref{fig:openbookqa flip rates,fig:sports flip rates,fig:sqa flip rates} present flip rates for OpenBookQA, Sports and StrategyQA datasets, respectively. They follow similar trends and patterns as flip rates for ARC-Challenge dataset presented in \Cref{fig:arc flip rates}.

\subsection{Examples of Fail Cases}
\label{sec:fail case example}
Since models often exhibit weak robust self-consistency under input perturbations and weak responsive self-consistency under CoT perturbations, we provide examples of such fail cases in \Cref{fig:example flip when input strength 0,fig:example not flip when strength 3}, respectively.

\begin{figure*}
    
    \sessiondivider{Session 1 (original answer and CoT)}
    \begin{tcolorbox}[humanbox]
    A student has \textbf{one liter of feathers and \textcolor{red}{one liter of iron nails}}. Which statement best compares the masses of the two samples? \\
    Options: \\
    A): The masses cannot be compared without a balance. \\
    B): The mass of the feathers is less than the mass of the nails \\
    C): The mass of the feathers is the same as the mass of the nails. \\
    D): The mass of the feathers is greater than the mass of the nails. 
    \end{tcolorbox}
    {\raggedright\originalquestion}
    \begin{tcolorbox}[assistantbox]
    The student has one liter of feathers and one liter of iron nails. \\
    The volume of both samples is the same (one liter). \\
    The mass of a substance is equal to its density multiplied by its volume. \\
    The density of feathers is much less than the density of iron nails. Therefore, the mass of the feathers is less than the mass of the nails.
    \end{tcolorbox}
    \begin{tcolorbox}[assistantbox]
    B
    \end{tcolorbox}
    {\raggedleft\originalanswer}
    \sessiondivider{Session 2 (answer after input perturbation)}
    \begin{tcolorbox}[humanbox]
    A student has \textbf{one liter of feathers and \textcolor{red}{two liters of iron nails}}. Which statement best compares the masses of the two samples?\\
    \texttt{[same answer options A-D]}
    \end{tcolorbox}
    {\raggedright\perturbedquestion}
    
    \begin{tcolorbox}[assistantbox]
    D
    \end{tcolorbox}
    {\raggedleft\newanswer}
    \captionof{figure}{Example of model changing answer under zero-strength input perturbation}
    \label{fig:example flip when input strength 0}
\end{figure*}

\Cref{fig:example flip when input strength 0} shows example of a model changing its answer under negligible input perturbation. In this example, the main reasoning path to get Option B still applies to the perturbed question, i.e.\ the perturbation does not interfere with the reasoning.
However, the model shows weak robustness to negligible perturbations and changes its answer.

\Cref{fig:example not flip when strength 3} shows example of a model not changing its answer under strong CoT perturbation. In this example, the original reasoning path states that the star ``exhausted its hydrogen'' and becomes ``cooler'' so the original answer is A. The changes in perturbed CoT substantially interfere with the original reasoning path because it now states that the star becomes ``hotter'', which removes support for answer A, and ``gained hydrogen'' which can shift support to answer D. However, we can see that the model's answer is still A given this perturbed CoT. 

\begin{figure*}
    \sessiondivider{Session 1 (original answer and CoT)}
    \begin{tcolorbox}[humanbox]
    A red giant star differs from main sequence stars because it \\
    Options: \\
    A): burns at a cooler temperature. \\
    B): is always larger than other suns. \\
    C): has greater density and gravity \\
    D): fuses hydrogen to make helium
    \end{tcolorbox}
    {\raggedright\originalquestion}
    \begin{tcolorbox}[assistantbox]
    A red giant star is a stage in the life of a star, just like the main sequence stage. \\
    The main difference is that \textbf{a red giant star has \textcolor{red}{exhausted its hydrogen fuel} and has expanded to become much larger and \textcolor{red}{cooler}} than it was during its main sequence life.
    \end{tcolorbox}
    {\raggedleft\originalcot}
    \begin{tcolorbox}[assistantbox]
    A
    \end{tcolorbox}
    {\raggedleft\originalanswer}
    \sessiondivider{Session 2 (answer after CoT perturbation)}
    \begin{tcolorbox}[humanbox]
    \texttt{[same question and answer options]}
    \end{tcolorbox}
    
    \begin{tcolorbox}[assistantbox]
    A red giant star is a stage in the life of a star, just like the main sequence stage. \\
    The main difference is that \textbf{a red giant star has \textcolor{red}{gained hydrogen fuel} and has shrunk to become much smaller and \textcolor{red}{hotter}} than it was during its main sequence life.
    \end{tcolorbox}
    {\raggedleft\perturbedcot}
    \begin{tcolorbox}[humanbox]
    The single, most likely answer is (
    \end{tcolorbox}
    \begin{tcolorbox}[assistantbox]
    A
    \end{tcolorbox}
    {\raggedleft\newanswer}
    \captionof{figure}{Example of model not changing answer under high-strength CoT perturbation}
    \label{fig:example not flip when strength 3}
\end{figure*}

\subsection{Direct Prompting vs CoT Prompting for Input Perturbation}
\label{sec:direct prompting vs cot prompting}
\begin{figure*}
    \centering
    \includegraphics[width=1\linewidth]{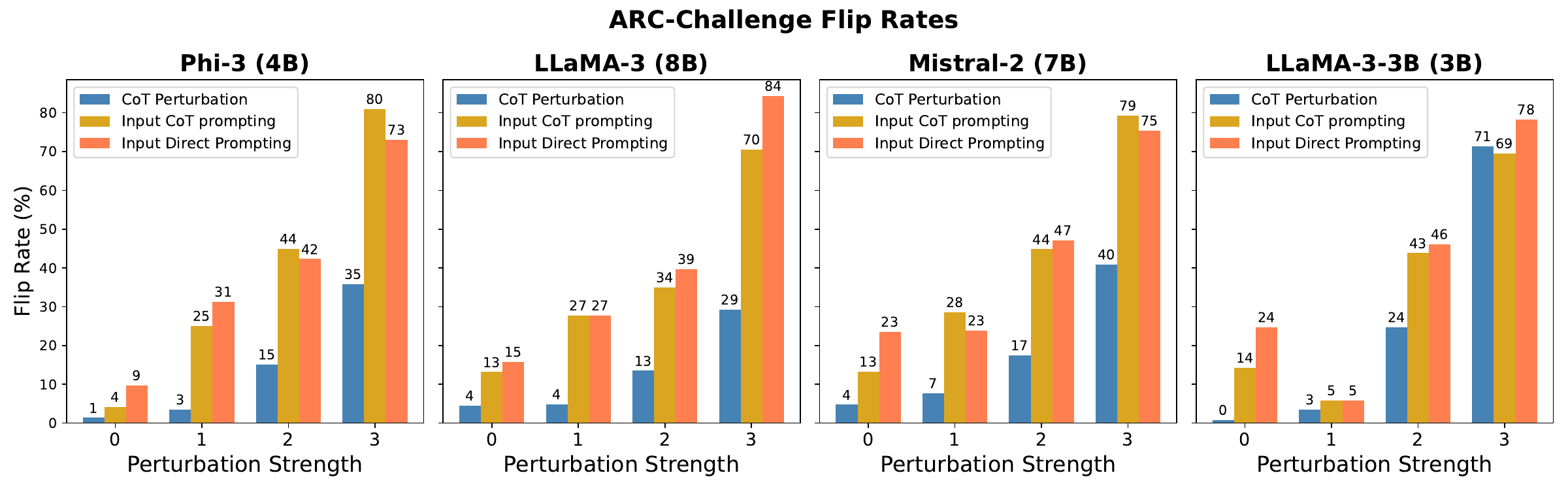}
    \caption{ARC-Challenge flip rates of four CoT models under 3 settings: CoT perturbation, input perturbation using direct prompting and input perturbation using CoT prompting}
    \label{fig:arc flip rates 3 bars}
\end{figure*}

\begin{figure*}
    \centering
    \includegraphics[width=1\linewidth]{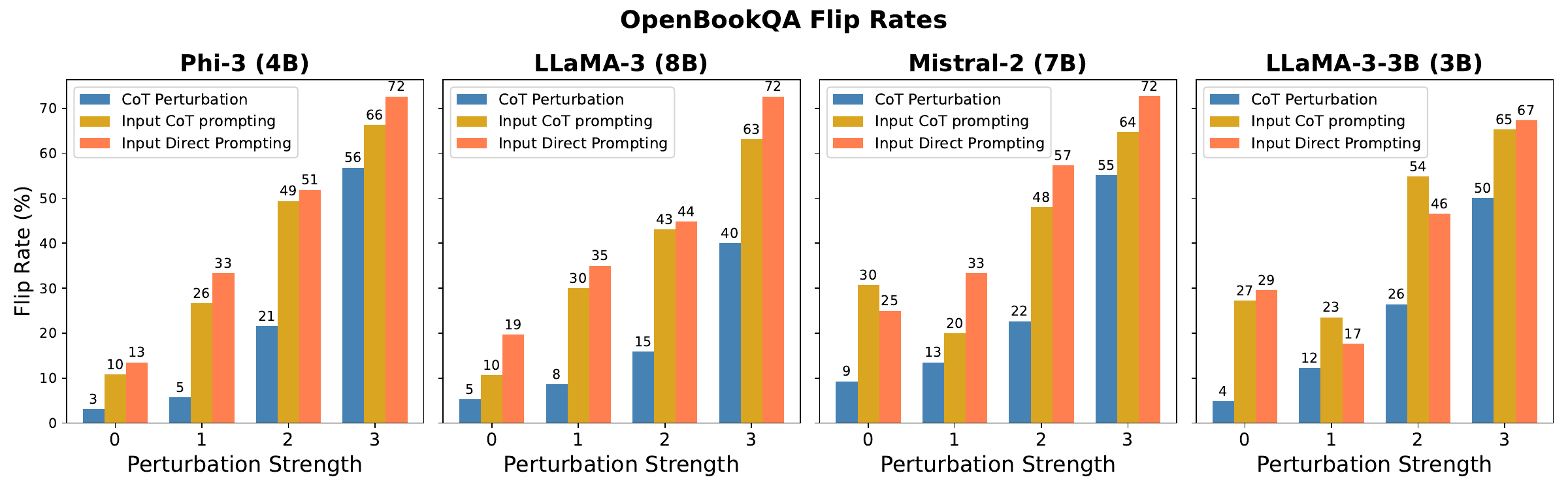}
    \caption{OpenBookQA flip rates of four CoT models under 3 settings: CoT perturbation, input perturbation using direct prompting and input perturbation using CoT prompting}
    \label{fig:openbookqa flip rates 3 bars}
\end{figure*}

\begin{figure*}
    \centering
    \includegraphics[width=1\linewidth]{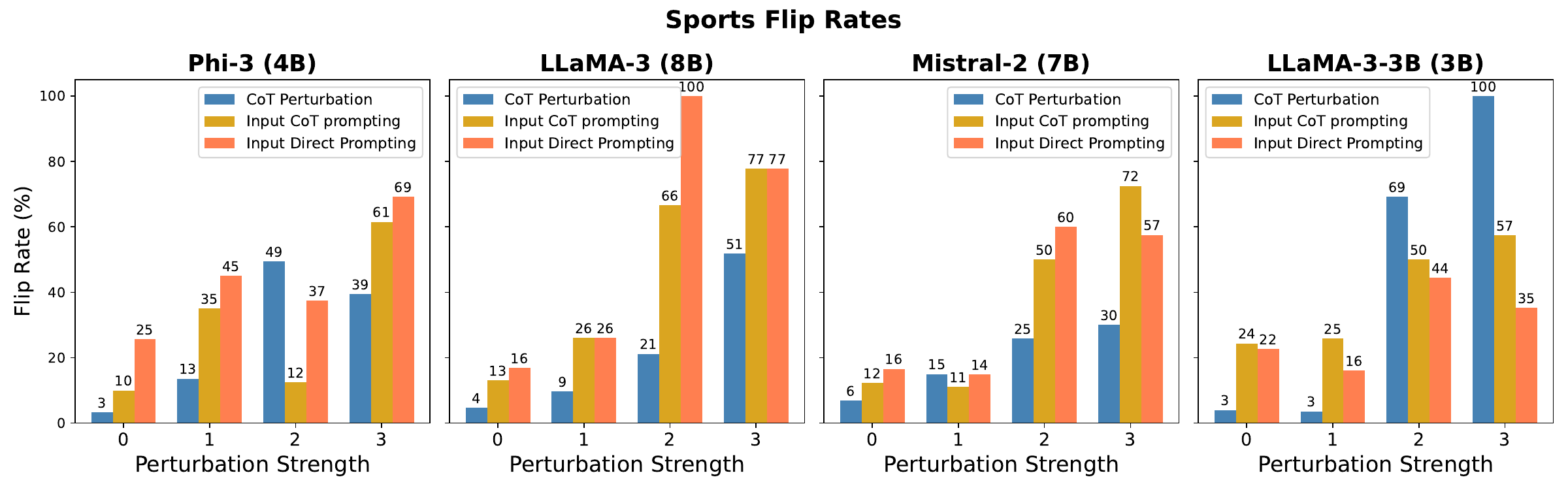}
    \caption{Sports flip rates of four CoT models under 3 settings: CoT perturbation, input perturbation using direct prompting and input perturbation using CoT prompting}
    \label{fig:sports flip rates 3 bars}
\end{figure*}

\begin{figure*}
    \centering
    \includegraphics[width=1\linewidth]{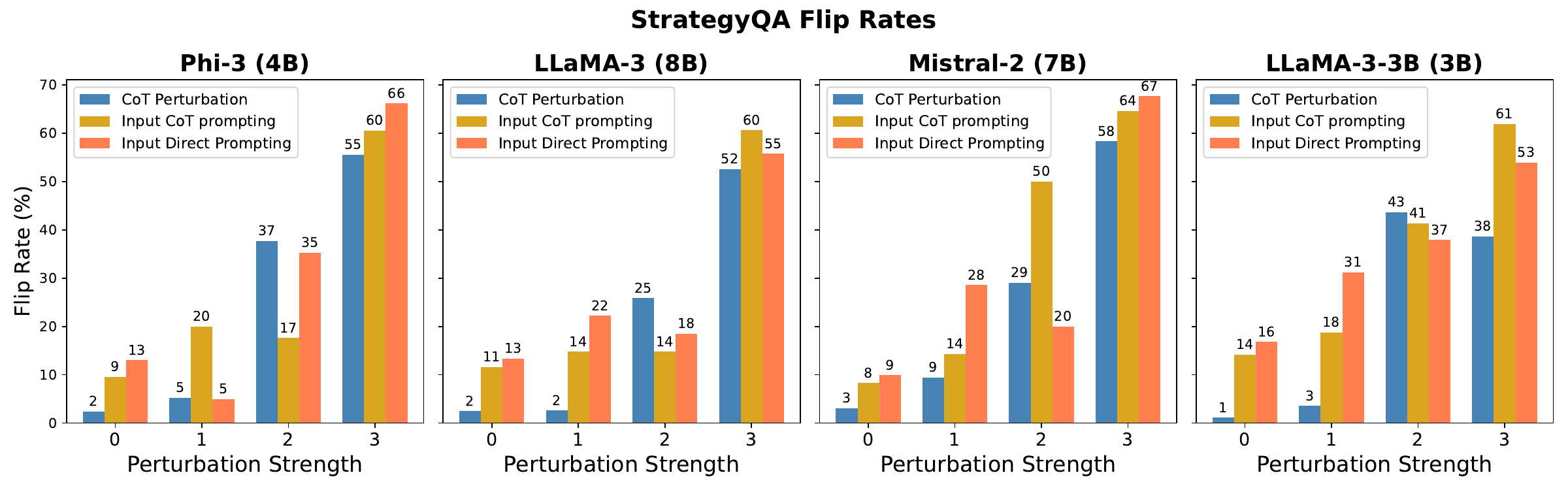}
    \caption{StrategyQA flip rates of four CoT models under 3 settings: CoT perturbation, input perturbation using direct prompting and input perturbation using CoT prompting}
    \label{fig:strategyqa flip rates 3 bars}
\end{figure*}

In addition to the direct prompting approach used to extract the model answers to perturbed questions, we also attempt to extract the answers using CoT prompting approach, which prompts the model to generate a CoT to the perturb question first, then concatenate the question and the generated CoT to trigger for the final answer. We then compute and plot the flip rates for each perturbation strength under three settings: CoT perturbation, input perturbation with direct prompting, and input perturbation with CoT prompting. The results are presented in \Cref{fig:arc flip rates 3 bars,fig:openbookqa flip rates 3 bars,fig:sports flip rates 3 bars,fig:strategyqa flip rates 3 bars}.

From \Cref{fig:arc flip rates 3 bars,fig:openbookqa flip rates 3 bars,fig:sports flip rates 3 bars,fig:strategyqa flip rates 3 bars}, we generally observe similar trends between input perturbations using direct prompting and CoT prompting. More specifically, input perturbation still generally have higher flip rates than CoT perturbation, regardless of whether we use direct prompting or CoT prompting. In addition, both prompting approaches also give similar flip rates. Since the result patterns of input perturbation compared to CoT perturbation are similar regardless of the prompting approach, we decide to focus our subsequent analysis only on the results of input perturbation using direct prompting, which minimises the extraneous variables as discussed in \Cref{sec:data input perturbation}.

\end{document}